\documentclass[journal,twoside,web]{ieeecolor}
\usepackage{generic}

\usepackage{cite}
\usepackage{amssymb}
\usepackage{latexsym}

\usepackage{url}
\usepackage{xcolor}

\usepackage{hyperref}
\usepackage{graphicx}
\usepackage{amsmath}
\usepackage{amssymb}
\usepackage{verbatim}
\usepackage{booktabs}
\usepackage{multirow}
\usepackage{bm}
\usepackage{array}
\usepackage{graphicx}
\usepackage{epstopdf}
\usepackage{subfigure}
\usepackage{amssymb}
\usepackage{amsmath}
\usepackage{makecell}

\def\0{{\bf 0}}
\def\1{{\bf 1}}

\def\etal{{\em et al.}}
\def\eg{{\em e.g.}}
\def\ie{{\em i.e.}}

\def\etal{{\em et al.\/}\,}

\usepackage{tabularx} 
\usepackage{booktabs}   
\usepackage{mathrsfs} 
\usepackage{multirow} 
\usepackage{amsfonts} 
\usepackage{wrapfig}  
\usepackage{cite}
\usepackage{footnote}

\begin{document}

\title{Keeping Medical AI Healthy and Trustworthy: A Review of Detection and Correction Methods for System Degradation
}

\author{Hao~Guan,  
\and David~Bates,
\and
    Li~Zhou
\thanks{H.~Guan, D.~Bates and L.~Zhou are with the Division of General Internal Medicine and Primary Care, Brigham and Women’s Hospital, and Department of Medicine, Harvard Medical School, Boston, Massachusetts 02115, USA.  
Corresponding Author: H.~Guan (hguan6@bwh.harvard.edu).} 
}
\maketitle

\begin{abstract}
Artificial intelligence (AI) is increasingly integrated into modern healthcare, offering powerful support for clinical decision-making. However, in real-world settings, AI systems may experience performance degradation over time, due to factors such as shifting data distributions, changes in patient characteristics, evolving clinical protocols, and variations in data quality. 
These factors can compromise model reliability, posing safety concerns and increasing the likelihood of inaccurate predictions or adverse outcomes. 
This review presents a forward-looking perspective on monitoring and maintaining the “health” of AI systems in healthcare. We highlight the urgent need for continuous performance monitoring, early degradation detection, and effective self-correction mechanisms. The paper begins by reviewing common causes of performance degradation at both data and model levels. We then summarize key techniques for detecting data and model drift, followed by an in-depth look at root cause analysis. Correction strategies are further reviewed, ranging from model retraining to test-time adaptation. Our survey spans both traditional machine learning models and state-of-the-art large language models (LLMs), offering insights into their strengths and limitations. Finally, we discuss ongoing technical challenges and propose future research directions. This work aims to guide the development of reliable, robust medical AI systems capable of sustaining safe, long-term deployment in dynamic clinical settings.

\end{abstract}

\begin{IEEEkeywords}
AI in Healthcare, AI Performance Degradation, AI Self-Correction, Data Shift, Model Drift, Clinical Decision Support, AI Monitoring, AI Reliability, AI Safety
\end{IEEEkeywords}

%
\IEEEpeerreviewmaketitle

\section{Introduction}

\IEEEPARstart{T}{he} world is seeing increasing applications of Artificial Intelligence (AI) in healthcare and medicine~\cite{yu2018artificial,moor2023foundation,li2024artificial}, transforming disease diagnosis, patient monitoring, outcome prediction, and treatment planning. AI technologies range from image analysis algorithms in radiology~\cite{hosny2018artificial} and pathology~\cite{bahadir2024artificial} to predictive models that leverage electronic health records (EHR) for early disease detection~\cite{guan2025cd}. 
By March 2025, the United States Food and Drug
Administration (FDA) has listed 1,016 AI-enabled medical devices that have met the FDA's applicable premarket requirements~\cite{FDA2024}, and the number of AI/ML-based algorithms approved for clinical use by FDA continues to grow at a rapid rate~\cite{joshi2024fda}.
As these AI systems continue to be adopted, ensuring their reliability and sustained performance becomes critically important.
A recent study~\cite{van2021artificial} analyzed 100 commercially available AI products in radiology and found that while AI adoption is accelerating, many systems lack rigorous scientific validation. Specifically, 64\% of the products had no peer-reviewed evidence, and only a small fraction demonstrated clinical impact beyond diagnostic accuracy. 
The reliability of AI models in healthcare is not merely a technological concern but a crucial factor in ensuring patient safety and quality of care. Inaccurate or degraded AI performance can lead to severe consequences, including misdiagnosis, inappropriate treatment recommendations, and a loss of trust from healthcare providers and patients. 

\begin{figure}[t]
\setlength{\belowcaptionskip}{-2pt}
\setlength{\abovecaptionskip}{0pt}
\setlength{\abovedisplayskip}{0pt}
\setlength{\belowdisplayskip}{0pt}
\center
 \includegraphics[width= 1.0\linewidth]{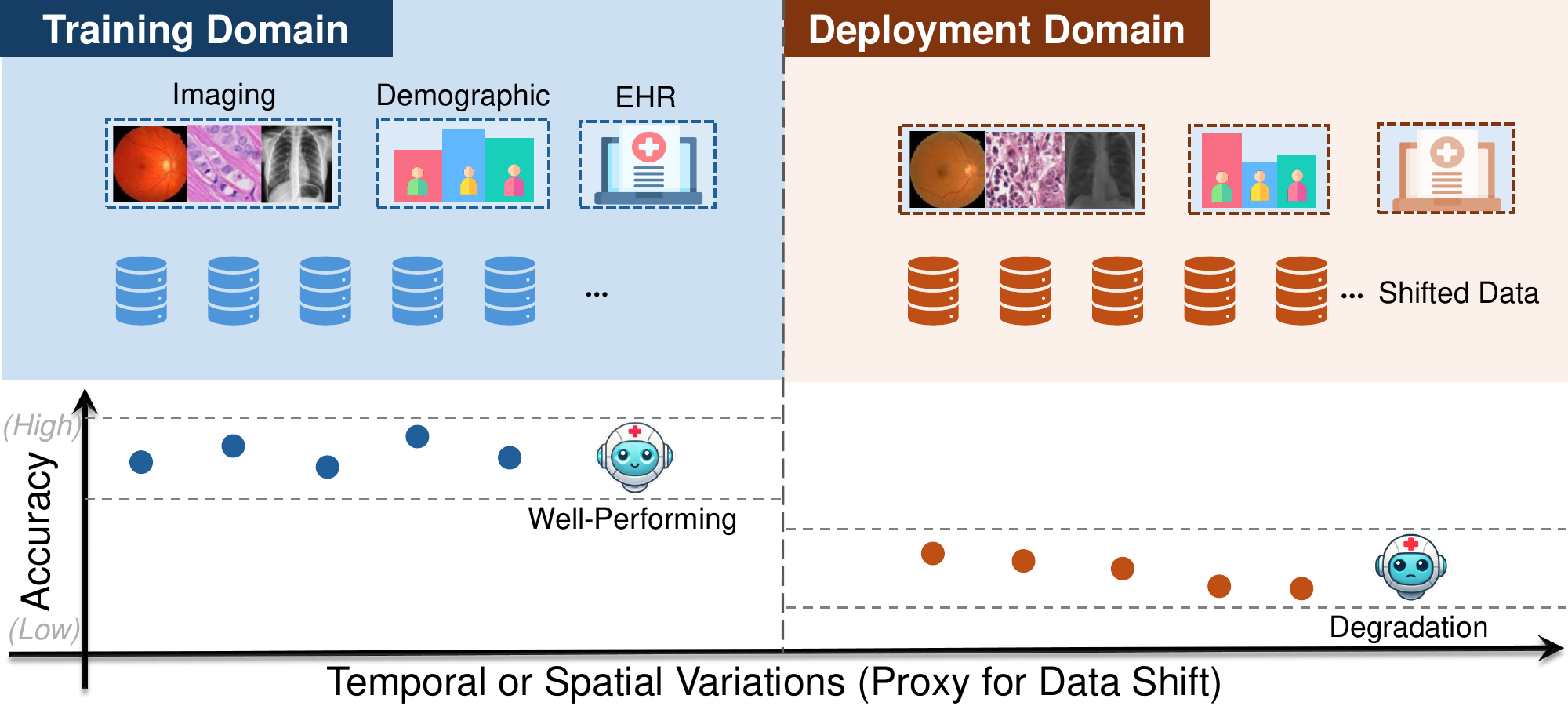}
 \caption{Illustration of performance degradation of medical AI models under temporal or spatial variations. The x-axis represents differences across time, or sites (proxy for data shift). Training data (left) include imaging, demographic, and EHR modalities, whereas deployment data (right) exhibit distributional changes (``shifted data"). Such shifts, regardless of their source, lead to reduced model accuracy and degraded performance, highlighting the importance of continuous AI monitoring.} 
 \label{Degradation}
\end{figure}

Performance degradation in AI, or model drift~\cite{thompson2024implementing,lacson2022machine}, as shown in Fig.~\ref{Degradation}, refers to the phenomenon where models exhibit reduced effectiveness in real-world applications compared to their performance during initial training or testing. 
In classic machine learning (ML) theory, training and test data are assumed to be drawn from the same underlying distribution~\cite{murphy2012machine}.
However, this cannot hold in real-world practice.
This issue is particularly pressing in healthcare, where models are exposed to dynamic, real-world data that may differ significantly from the training data.
Generally, these data variations that lead to model degradation/corruption can be categorized into two main types. The first is cross-environment variation, also referred to as \textbf{spatial or location-based variation}. An AI model trained in one environment, when applied to new environments, may encounter significantly different data distributions. For example, demographic differences in a new clinical setting may contrast sharply with the data distributions in the original training environment, leading to performance degradation. Differences in medical practices, device settings, or even disease prevalence across geographic regions further exacerbate this issue.

Another type of variation is within-environment variation, also known as \textbf{temporal variation}. In this scenario, an AI model is designed and deployed within a fixed environment, meaning the external setting remains constant. However, over time, the model still experiences performance degradation. This phenomenon can be vividly depicted as “AI aging”~\cite{vela2022temporal}.
Please note this degradation arises from evolving variations in the underlying data distribution rather than time itself as a causal factor. Here, time serves as a proxy for data or model drift. 
Such temporal variations can occur abruptly, triggering a sharp drop in performance that can be classified as an anomaly event. Conversely, a more subtle but persistent issue arises when the AI model suffers from a gradual performance decline, which may go unnoticed without continuous monitoring.

Only recently has the research community and industry started recognizing the importance of monitoring AI performance degradation~\cite {vokinger2021continual,FDA2019,feng2022clinical}.
Chen~\etal\cite{chen2017decaying} demonstrate that clinical decision support systems trained on historical EHR data experience significant declines in predictive accuracy as the underlying clinical practice evolves, suggesting a natural decay in the utility of past clinical data. Their findings show that AI models relying on older datasets performed worse than those trained on more recent data, indicating an inherent susceptibility to performance degradation. 

The study~\cite{young2022empirical} empirically evaluates the long-term performance degradation of ML models predicting in-hospital mortality at the time of emergency admission, using a real-world dataset of 1.83 million patient discharge records from Maryland State Inpatient Database (2016–2018). By analyzing four top-performing ML models over 2.5 years, the authors track degradation across key metrics, including Area Under the Receiver Operator Characteristic Curve (AUROC), accuracy, precision, and recall. The findings reveal that while ML models can remain effective for over a year post-training, gradual performance decline necessitates strategic retraining. Additionally, the study highlights the limitations of 10-fold cross-validation in predicting long-term reliability and emphasizes the need for adaptive monitoring to maintain robustness in clinical AI applications.

The study~\cite{dong2024performance} investigates performance drift in machine learning models used for cardiac surgery risk prediction, using a large UK cardiac surgery dataset from 2012–2019. They develop and evaluate five ML models (\eg, XGBoost, Random Forest) and the findings highlight strong evidence of performance degradation over time, attributed to changes in data distributions and variable importance. 

Another research on models predicting acute kidney injury (AKI)~\cite{davis2017calibration} over 9 years has shown that while discriminatory performance (\eg, ranking outcomes) may remain stable, calibration drift (the misalignment of predicted probabilities with actual outcomes) can significantly undermine their clinical utility. 
Similar findings are reported in~\cite{davis2018calibration}, where multiple regression and machine learning models exhibit calibration drift in the task of predicting 30-day hospital mortality.
These emphasize the importance of performance monitoring to detect and correct degradation, ensuring AI models maintain their performance in dynamic healthcare environments.

Even the most advanced AI models are not immune to unexpected performance degradation in real-world applications. The rise of large language models (LLMs) has marked a significant breakthrough in AI research, attracting global attention for their impressive capabilities~\cite{thirunavukarasu2023large}. However, the increasing integration of LLMs into medicine and clinical workflows highlights the urgent need for robust AI performance monitoring. A recent study~\cite{payne2024performance} assesses GPT-4's performance on the American college of Radiology in-training examination and finds that when the questions were repeated several months later to assess for model drift, GPT-4 chose a different answer 25.5\% of the time, indicating substantial temporal variability in LLM performance.  
A similar trend is observed in radiology question answering and interpretation tasks, where GPT-4 also demonstrates performance degradation over time~\cite{bhayana2023performance}. 
A broader evaluation~\cite{chen2024chatgpt} highlights how GPT-4 and GPT-3.5 exhibit performance degradation across various tasks, including code generation, math problem-solving, and sensitive question answering. One key observation is that GPT-4's ability to follow user instructions degrades over time, contributing to the observed performance drops.
These studies underscore the fact that even the most advanced AI systems can exhibit unexpected performance degradation in real-world scenarios across both clinical and non-clinical domains, reinforcing the necessity for continuous monitoring to ensure long-term reliability and safety.

\begin{figure}[t]
\setlength{\belowcaptionskip}{-2pt}
\setlength{\abovecaptionskip}{0pt}
\setlength{\abovedisplayskip}{0pt}
\setlength{\belowdisplayskip}{0pt}
\center
 \includegraphics[width= 1.0\linewidth]{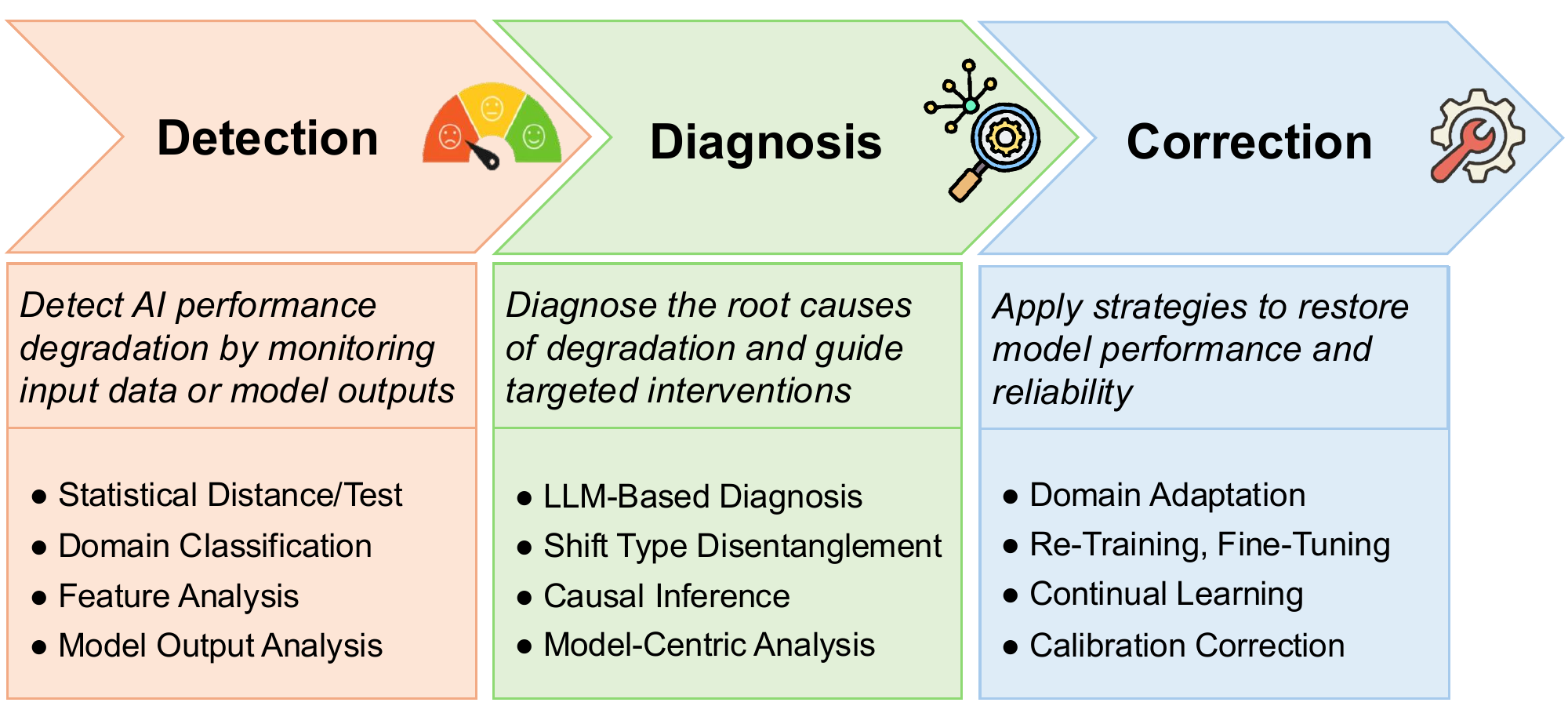}
 \caption{Overview of the medical AI monitoring framework, from performance degradation detection, diagnosis to correction.} 
 \label{Main}
\end{figure}

Our study is motivated by the growing use of AI in the medical and healthcare fields. As these systems are increasingly applied to real-world clinical settings, concerns about their long-term safety and reliability are becoming more important~\cite{albahri2023systematic}. In particular, there is a growing awareness that monitoring AI after deployment deserves more attention~\cite{rajagopal2024machine,khattak2023mlhops}.
However, despite this need, there is still no comprehensive review that looks at the full picture of AI monitoring in medicine and healthcare, from detecting performance issues, to understanding their causes, to applying correction methods.

To address this gap, our survey organizes existing methods for detecting, diagnosing, and correcting AI performance degradation in healthcare within a unified Detection-Diagnosis-Correction (DDC) cycle (Fig.~\ref{Main}). The DDC framework aligns closely with the \textbf{FDA's Predetermined Change Control Plan (PCCP)}~\cite{FDA_PCCP} for AI-enabled device software: ``Detection" provides monitoring signals and predefined triggers outlined in the ``Modification Protocol", grounded in PCCP's ``Data Management Practices" and ``Performance Evaluation" components; ``Diagnosis" corresponds to ``Impact Assessment" and ``Traceability", linking each proposed modification to its supporting protocol elements; ``Correction" involves executing ``Re-Training Practices", verifying outcomes through performance evaluation, and implementing ``Update Procedures". In parallel, the DDC framework supports \textbf{ISO/IEC 42001}~\cite{ISO_IEC_42001_2023} audit trails by reinforcing lifecycle documentation, decision transparency, and risk control. Collectively, this structure is designed to help medical AI systems remain safe, effective, and trustworthy throughout their deployment lifecycle.


\section{Problem Formulation}

Ensuring the long-term reliability and robustness of medical AI models requires more than pre-deployment evaluation on training and test sets. To avoid ambiguity, we distinguish between these related but distinct concepts:
\begin{itemize}
\item \textbf{Reliability} refers to the ability of a medical AI system to consistently deliver accurate and trustworthy performance across time, populations, and clinical settings, ensuring predictable behavior throughout its lifecycle.
\item \textbf{Robustness} refers to the ability of a model to maintain stable performance under unexpected perturbations, such as noisy or corrupted inputs, and distributional shifts.
\end{itemize}

In this survey, we view robustness as a necessary component of reliability: a robust model is more likely to be reliable, but reliability additionally requires continuous monitoring, calibration, and correction over time. 
Building on this foundation, we define AI performance monitoring as comprising two key components: \textbf{data monitoring} and \textbf{model monitoring}, both framed as distributional comparisons between datasets or model behaviors. 
To avoid ambiguity, we distinguish these two components. Data monitoring focuses on identifying distributional changes in the input space (\eg, covariate shift), which may eventually lead to degraded model performance. In contrast, model monitoring targets performance decay or behavioral drift of the model itself. Importantly, model drift may stem from, but is not limited to, data drift. It can also arise from architectural or optimization-related factors. This separation ensures that our taxonomy captures both external changes in data and internal changes in model behavior.


\subsection{Data Monitoring}

Let \( \mathcal{X} \subset \mathbb{R}^d \) denote the input space. Suppose we are given two datasets:
\begin{equation}
   D_1 = \{x_i^{(1)}\}_{i=1}^{n_1} \sim P_1(X), \quad D_2 = \{x_j^{(2)}\}_{j=1}^{n_2} \sim P_2(X) 
\end{equation}
The goal of \textit{data monitoring} is to detect whether a distributional shift has occurred, that is,
\begin{equation}
H_0: P_1(X) = P_2(X) \quad \text{vs.} \quad H_1: P_1(X) \neq P_2(X)
\end{equation}
This is done by computing a statistical discrepancy measure \( \delta(P_1, P_2) \). A shift is detected when:
\begin{equation}
\delta(P_1, P_2) > \epsilon
\end{equation}
for some predefined threshold \( \epsilon \).
This formulation applies to two common settings. 

In the \textbf{{static setting}}, also referred to as between-site shift, \( D_1 \) and \( D_2 \) are fixed datasets collected from different sources such as hospitals, imaging devices, or population groups. The goal is to assess across-domain generalization through a one-time comparison.

In the \textbf{{temporal setting}}, new data arrive as a stream \( \{x_t\}_{t=1}^{\infty} \), and drift is monitored continuously by comparing the current distribution $P_{\text{cur}}$ (data in a more recent window: $D_{\text{cur}} = \{x_t\}_{t=t_1}^{t_1 + w}$) with a reference distribution $P_{\text{ref}}$. 
The comparison remains the same:
\begin{equation}
\delta(P_{\text{ref}}, P_{\text{cur}}) > \epsilon
\end{equation}
Both settings are mathematically equivalent and involve comparing the distributions of two sample sets.
\subsection{Model Monitoring}
Let \( f: \mathcal{X} \rightarrow \mathcal{Y} \) be a deployed model that maps inputs to predictions. Given a stream of input data \( \{x_t\} \), we denote the corresponding model outputs as \( \hat{y}_t = f(x_t) \). Let \( O_t = \{\hat{y}_i\}_{i \in t} \) denote the set of predictions made within a given time window \( t \).
\textit{Model monitoring} aims to detect whether the model’s behavior has changed over time, either in terms of its predictive accuracy or the statistical properties of its outputs.

\textbf{{Supervised model monitoring}}: applies when ground-truth labels \( y_t \) are available. Let \( D_t = \{(x_i, y_i)\}_{i \in t} \) be the labeled dataset at time \( t \), and define a performance metric \( m(f; D_t) \), such as accuracy, AUROC or F1 score. Model degradation is detected when:
\begin{equation}
m(f; D_t) < m(f; D_{\text{val}}) - \delta_m
\end{equation}
where \( D_{\text{val}} \) is the validation dataset used as a baseline, and \( \delta_m \) is a predefined degradation threshold.

\textbf{{Unsupervised model monitoring}}: applies when ground-truth labels are not available. In this case, we compare the output distribution \( P(O_t) \) at time \( t \) with a reference output distribution \( P(O_{t_0}) \) from an earlier time \( t_0 \), using a divergence function \( \delta \). A model shift is detected if:
\begin{equation}
\delta(P(O_t), P(O_{t_0})) > \epsilon
\end{equation}
We can monitor the entropy of the model’s softmax outputs, 
as well as other statistics such as output confidence, class-wise prediction frequencies, and prediction stability.


\subsection{Operational Parameters Selection for Drift Detection}
Drift detection typically requires setting two key parameters: a threshold and a window size.
The threshold (\eg, $\epsilon$ for data shift in (3) or $\delta_m$ for model drift in (5)) defines when the difference between current and reference data is large enough to be considered meaningful. In practice, even in post-deployment scenarios where labels are delayed or unavailable, it can be determined from prior experiments or adjusted to maintain the false-alarm rate within an acceptable range (\eg, 1-5\%). Thresholds can also be refined adaptively once delayed ground-truth labels become available.

The window size (w) determines how much data is compared at each monitoring step. A smaller window increases sensitivity to short-term changes, whereas a larger window improves stability by reducing noise. In practice, rolling windows containing hundreds of samples or data from the past 1-3 months are commonly used, although the optimal choice ultimately depends on the specific system and task.


\section{Data-Driven Causes of Medical AI Degradation}

\subsection{Data Shift in Medical AI}
In disease diagnosis, AI models typically use biomedical features (\eg, laboratory test results, vital signs, or radiomic features) to classify patients as either disease-positive or disease-negative. However, in real-world deployment, various types of data shifts can degrade performance~\cite{finlayson2021clinician}. Let $\mathbf{X}$ denote input features and $\mathbf{Y}$ the target output; data shift refers to changes in their joint distribution $P(\mathbf{X}, \mathbf{Y})$. These changes can arise from shifts in $P(\mathbf{X})$ (covariate shift), $P(\mathbf{Y})$ (label shift or prior probability shift), or $P(\mathbf{Y}|\mathbf{X})$ (concept or relationship shift)~\cite{moreno2012unifying}.

\begin{figure}[t]
\setlength{\belowcaptionskip}{0pt}
\setlength{\abovecaptionskip}{0pt}
\setlength{\abovedisplayskip}{0pt}
\setlength{\belowdisplayskip}{0pt}
\center
 \includegraphics[width= 1.0\linewidth]{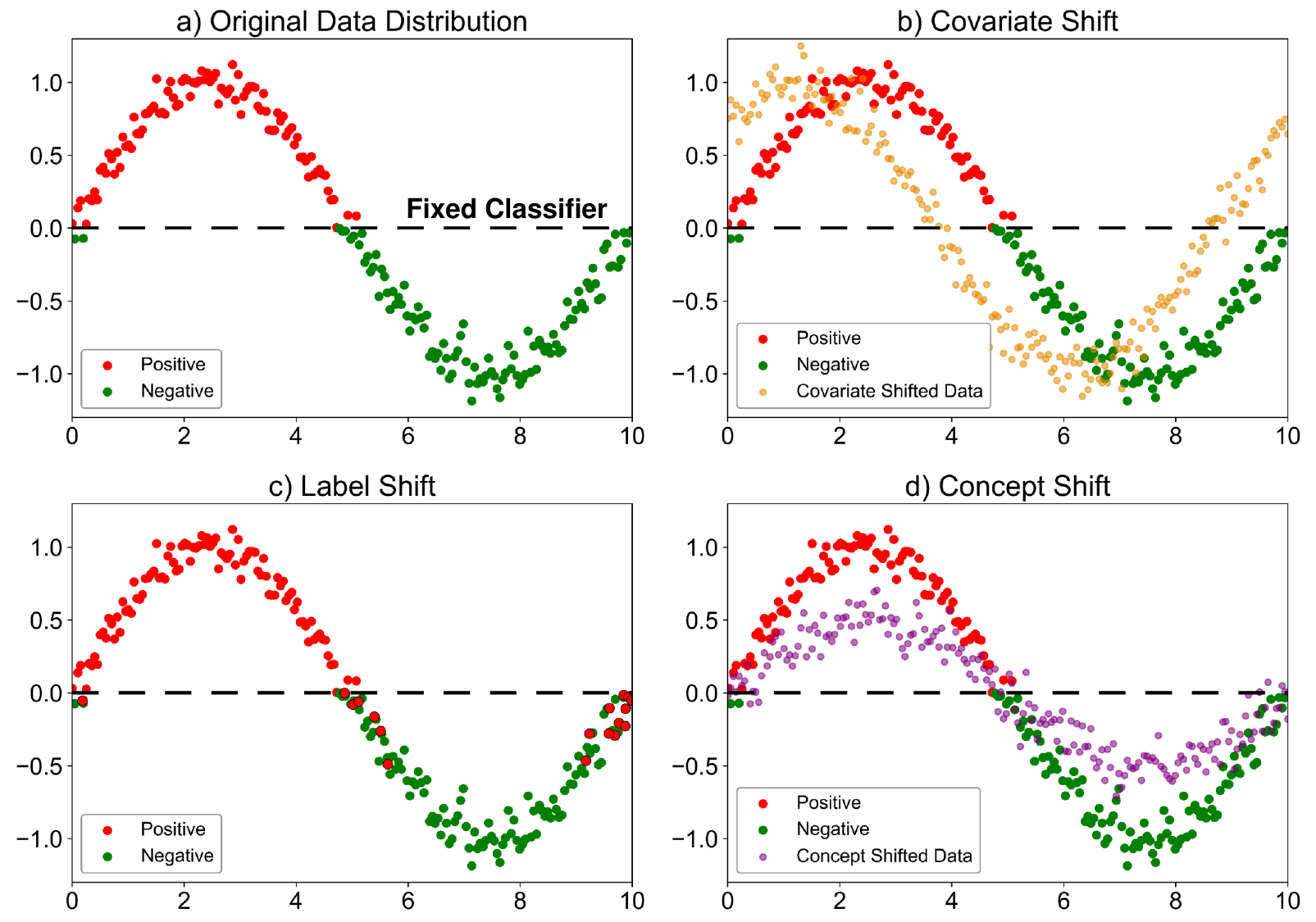}
 \caption{Types of data shift in a disease classification model.
The model classifies positive (disease, red) and negative (normal, green) cases using a fixed classifier (dashed line).
a) Original data distribution.
b) Covariate shift: changes in the input feature distribution (\eg, due to imaging acquisition differences) while the classifier remains fixed.
c) Label shift: changes in class priors (\eg, increase in disease prevalence) under a fixed classifier.
d) Concept shift: changes in the relationship between features and target output.} 
 \label{data-shift}
\end{figure}

Fig.~\ref{data-shift} illustrates three types of data shift using synthetic data. 
We denote the source and target domains (or time windows) as $\mathcal{D}_s = (X_s, Y_s)$ and $\mathcal{D}_t = (X_t, Y_t)$, respectively. The distributions in the source and target are represented by $P_s(\cdot)$ and $P_t(\cdot)$. The data shifts are categorized based on what changes and what remains invariant across domains.

\subsubsection{Covariate Shift}
Refers to the case where the marginal distribution of the input changes, but the conditional distribution of labels given input remains the same:
\begin{equation}
P_s(X) \neq P_t(X), \quad \text{while} \quad P_s(Y \mid X) = P_t(Y \mid X)
\end{equation}

\textbf{Example:}  
The model is trained on data from an elderly population but deployed in a younger cohort, with consistent diagnostic logic.
\subsubsection{Label Shift}
Occurs when the marginal distribution of labels changes, while the class-conditional input distribution remains invariant:
\begin{equation}
P_s(Y) \neq P_t(Y), \quad \text{while} \quad P_s(X \mid Y) = P_t(X \mid Y)
\end{equation}

\textbf{Example:}  
A disease classification model is trained on data with low disease prevalence, but deployed during an outbreak where the prevalence significantly increases.

\subsubsection{Concept Shift}

Concept shift occurs when the conditional distribution of labels given inputs changes, potentially alongside changes in the input distribution in practice:
\begin{equation}
P_s(Y \mid X) \neq P_t(Y \mid X), \quad \text{while} \quad P_s(X) = P_t(X)
\end{equation}

\textbf{Example:}  
Updated clinical guidelines may lead to different labels for the same patient features.

\subsubsection{Case Studies}
The study in~\cite{rahmani2023assessing} explores the impact of data shift on machine learning models deployed in healthcare. Through simulations, it investigates covariate shift, concept shift, and major events (\ie, the COVID-19 pandemic) as sources of data shift. The study highlights that data shift significantly impacts model performance, necessitating monitoring and retraining to maintain effectiveness.

\subsection{Data Anomaly in Medical AI}

\subsubsection{Soft Failure vs. Hard Failure}
In contrast to data shifts, which represent \textbf{distribution-level} changes, data anomalies refer to \textbf{instance-level} irregularities in input data that can cause sudden and severe model failure. These anomalies occur at the point level and are not necessarily associated with global distributional trends.

Data shift is considered as a ``soft failure'' factor as models typically experience a gradual performance degradation due to slow changes in input distribution over time.
In contrast, data anomalies are often sudden and severe, acting as ``hard failure'' triggers where even a single anomalous input can lead to unpredictable or harmful model behavior.

\subsubsection{Definition and Types of Data Anomalies}

Let $x^*$ be an input from the current environment. A data anomaly is defined as:
\begin{equation}
x^* \sim P_t(X), \quad \text{but } x^* \notin \text{support}(P_s(X))
\end{equation}
where $P_s(X)$ is the distribution of the training data and $P_t(X)$ is the current test distribution. Although $x^*$ may technically belong to the test domain, it lies far outside the support of the training distribution and thus cannot be reliably handled by the model.

Data anomalies encompass a wide variety of problematic inputs, including but not limited to:
1) \textbf{Highly Noisy Inputs:} Corrupted or low-quality data that reduce signal-to-noise ratio;
2) \textbf{Missing or Incomplete Data:} Input with absent features, null values, or default placeholders;
3) \textbf{Highly Biased Inputs:} Data points that are unrepresentative or imbalanced, such as overrepresentation from a specific group, condition, or outlier case;
4) \textbf{Adversarial Inputs:} Carefully crafted samples designed to fool the model.

\subsubsection{Impact on Model Performance}
Data anomalies can cause sudden and serious model failures, unlike distribution shifts that usually lead to gradual performance drops (\eg, lower AUC). Even one bad input, like an extreme lab result or unreadable image, can lead to confident but wrong predictions, risking harmful clinical decisions.

While shift-based failures develop gradually and are often difficult to detect, sudden anomaly-induced errors are equally important. Robust AI monitoring should address both gradual distributional drift and abrupt data anomalies.


\section{Model-Driven Causes of Medical AI Degradation}
\subsection{Calibration Drift}

Calibration assesses how well predicted probabilities match actual outcomes~\cite{alba2017discrimination}. In a well-calibrated model, a prediction of 0.8 corresponds to an 80\% observed event rate. 
In healthcare, calibration is crucial for ensuring that predicted risks are both interpretable and actionable in clinical decision-making. 
However, when a model is deployed in new settings with different populations or disease prevalence, calibration can drift. This can distort risk estimates even when the model’s discrimination remains unchanged, undermining reliability and leading to suboptimal or inequitable clinical decisions.

Such model drift is especially critical in tasks like risk stratification, triage, and prognosis~\cite{davis2017calibration}, where miscalibrated predictions can misguide clinical decisions and erode trust. For example, a model trained in Hospital A with a 10\% sepsis rate may underpredict risk when deployed in Hospital B with a 20\% prevalence.
The study~\cite{davis2017calibration} explores seven acute kidney injury models over nine years and finds that while discrimination remained stable, calibration degraded over time due to shifts in event rates (label shift) and predictor-outcome relationships (concept shift), compromising risk estimates and decision support reliability.
\subsection{Model Optimization Trade-Offs}
Optimizing models for specific metrics, such as inference speed, computational efficiency, or adherence to safety constraints, can inadvertently degrade performance in other areas. For instance, the study by~\cite{payne2024performance} suggests that GPT-4's performance drift in radiology diagnosis tasks may be linked to optimization on other competing metrics, which potentially compromises diagnostic accuracy. These trade-offs can lead to unintended consequences in clinical AI models where accuracy and reliability are paramount.

A broader study~\cite{chen2024chatgpt} demonstrates GPT-4's performance decline across multiple tasks over time, underscoring the inherent difficulty of maintaining consistent performance. 
It highlights that fine-tuning or system update aimed at improving specific capabilities can cause unintended performance declines in other tasks.


\subsection{Catastrophic Forgetting}
To ensure adaptability in dynamic clinical environments, medical AI models are often updated with new patient data, clinical practices, or evolving guidelines~\cite{meijerink2024updating}. However, such update can unintentionally lead to catastrophic forgetting~\cite{kirkpatrick2017overcoming,li-etal-2024-revisiting}, where the model loses performance on previously learned knowledge. Importantly, this does not only affects past cases but also future cases that share highly similar patterns with those older examples. In high-stakes settings like healthcare, catastrophic forgetting of previously learned disease subtypes, rare conditions, or specific population subgroups can create critical blind spots with potentially harmful consequences. Mitigating this risk requires careful update strategies, such as rehearsal-based methods, regularization techniques, or freezing critical model components~\cite{perkonigg2021dynamic}.
\subsection{Knowledge Staleness}
Knowledge staleness occurs when models are not regularly updated, leading to a gradual mismatch between the model’s internal representations and current medical guidelines, clinical practices, or scientific discoveries. In healthcare, this can cause misclassifications, such as applying outdated diagnostic criteria or missing emerging diseases like COVID-19.
This problem is especially pronounced in LLMs~\cite{zhang-etal-2023-large,kandpal2023large}. Without timely updates, these models may produce hallucinations \cite{shah2024accuracy,pal2023med}, which are confident but incorrect outputs based on outdated or fabricated information.

To address knowledge staleness, medical AI systems should incorporate mechanisms for continual knowledge refresh. Common approaches include periodic retraining \cite{zheng2025towards}, retrieval-augmented generation (RAG) \cite{chen2024benchmarking}, and hybrid systems that combine LLMs with real-time databases \cite{peng2023check}.

\subsection{Prompt Sensitivity}
For models like LLMs, few-shot learning has demonstrated impressive performance in adapting to new tasks with minimal examples. However, studies~\cite{jeong2024limited} reveal that some LLMs in medical domain can suffer performance degradation after few-shot prompting, raising concerns about their robustness. Additionally, prompt sensitivity~\cite{sclar2024quantifying}, where variations in prompts lead to significantly different outputs, can result in inconsistencies and unpredictable behavior over time. Recent studies~\cite{sclar2024quantifying,zhuo2024prosa,razavi2025benchmarking} further show that current LLMs often struggle to fully capture and manage this sensitivity. This limitation is particularly critical in healthcare, where reliability, reproducibility, and consistency are paramount.
\section{Monitoring AI: Data Shift Detection}

\begin{figure}[t]
\setlength{\belowcaptionskip}{0pt}
\setlength{\abovecaptionskip}{0pt}
\setlength{\abovedisplayskip}{0pt}
\setlength{\belowdisplayskip}{0pt}
\center
 \includegraphics[width= 1.0\linewidth]{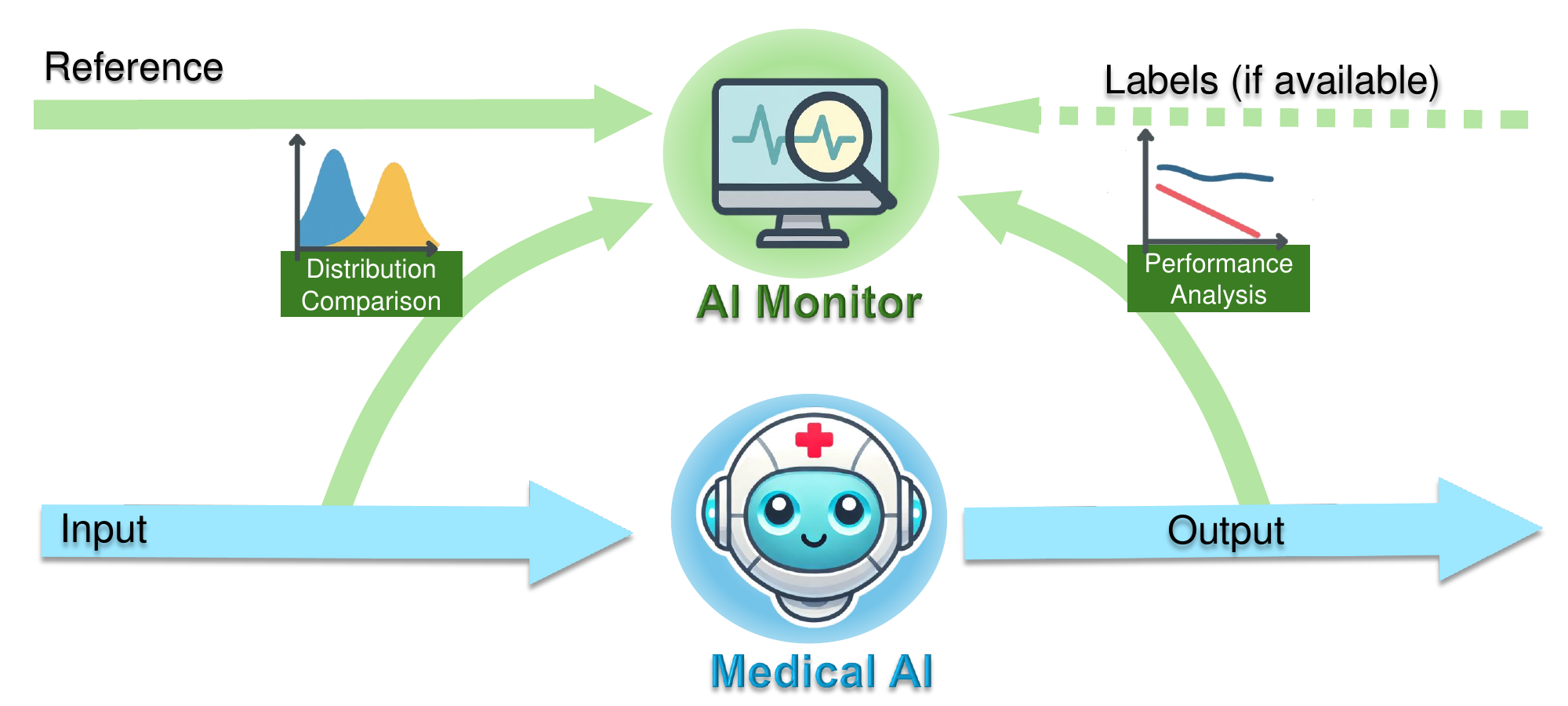}
 \caption{Illustration of AI performance monitoring: input (data) shift detection and output (model) drift detection.} 
 \label{Detection}
\end{figure}

In medical AI systems, performance degradation may arise from changes in input data (external) or the model itself (internal). Monitoring should therefore address both inputs and outputs, namely, input monitoring for data shift and output monitoring for model drift (Fig.~\ref{Detection}). This section focuses on data shift detection, a primary source of performance decline~\cite{sahiner2023data,stacke2020measuring}. Early detection enables timely tracking and maintenance of model performance.
\subsection{Statistic-Based Methods}

These methods detect distribution differences between datasets, whether across domains (spatial shift) or over time (temporal drift), by comparing feature representations from source and target data. They are valued for their statistical rigor, non-parametric nature, and broad applicability after feature extraction.
In addition, their outputs, such as distance scores or \emph{p}-values, are often interpretable, making them well-suited for explainable and reliable AI monitoring.
\subsubsection{Distance-Based Methods}
These methods detect distribution shift by measuring the discrepancy between feature distributions of source and target data. The data are first represented as feature vectors (raw, handcrafted, or model-generated) and the distance between distributions is computed in feature space. A larger distance suggests a higher likelihood of shift, typically assessed using a threshold or statistical test.

\vspace{5pt}

\noindent \textbf{Maximum Mean Discrepancy (MMD).}
MMD is a non-parametric distance metric used to compare two probability distributions based on their samples. It measures the difference between the mean embeddings of the source and target distributions in a reproducing kernel Hilbert space (RKHS). A larger MMD value indicates a greater discrepancy between the two distributions. MMD is especially well-suited for high-dimensional data.
In the study~\cite{kore2024empirical}, the authors utilize a pre-trained autoencoder to extract latent features from chest X-ray images in both source and target datasets. Then they apply MMD to compare the latent feature distributions between the datasets. A significant MMD score signals data shift such as those introduced by COVID-19.
In~\cite{dreiseitl2022comparison}, a Gaussian kernel-based MMD is used for detecting covariant shift of medical datasets and shows good performance.
\vspace{5pt}

\noindent \textbf{Wasserstein Distance.}
Also known as Earth Mover’s Distance, it quantifies the difference between two probability distributions by computing the minimum cost required to shift probability mass from one to the other. The cost reflects both how much mass is moved and how far it is moved, capturing the overall effort to align the distributions.

The study~\cite{stacke2020measuring} proposes a statistical method to detect domain shifts in medical imaging (\eg, pathology) for deep learning models. Features are first extracted from pathology images using a CNN, and shifts are quantified by comparing feature distributions using Wasserstein Distance. Validated on datasets like CAMELYON17, the method shows a strong link between data shift and model performance degradation.
\vspace{5pt}

\noindent \textbf{Kullback-Leibler (KL) Divergence.}
KL Divergence is a fundamental concept from information theory that measures how one probability distribution differs from another reference distribution. Specifically, it quantifies the expected extra information needed to represent data from the target distribution using a model trained on the source distribution. In simple terms, it reflects how inefficient it is to assume the data follows one distribution when it actually follows another.
The study~\cite{stacke2020measuring} uses KL Divergence to detect data shift in medical imaging datasets (\eg, histopathology).
\vspace{5pt}

\noindent \textbf{Jensen-Shannon (JS) Divergence.}
It is a symmetric and smoothed variant of KL divergence. It measures the similarity between two probability distributions.
In data shift detection, JS Divergence is commonly used to compare feature or prediction distributions between a source (training) and target (test) dataset. 
In a study analyzing temporal shifts in healthcare data~\cite{paiva2024new}, JS divergence is used to compare data shift over time in two datasets: MIMIC-IV and a Brazilian COVID-19 registry. The result shows its robustness for monitoring temporal drift in real-world healthcare data.
\subsubsection{Hypothesis Test Methods}
These methods assess whether two samples come from the same distribution using formal statistical tests. They are typically applied to low-dimensional or 1-D data where raw distributional differences are directly measurable. These methods output \emph{p}-values, enabling principled, threshold-free decisions. Their main strengths are simplicity, interpretability, and strong theoretical grounding, making them effective when assumptions are met.
\vspace{5pt}

\noindent \textbf{Kolmogorov-Smirnov (KS) Test.}
KS test is a non-parametric method for comparing two 1-D continuous distributions by evaluating the maximum difference between their cumulative distribution functions. It outputs a test statistic and a \emph{p}-value to assess whether the distributions differ significantly.
The CheXstray framework~\cite{merkow2023chexstray} applies KS test to monitor real-time data shift in healthcare by comparing DICOM metadata, image features, and model outputs to a reference dataset.
McUDI~\cite{poenaru2024mcudi} is an unsupervised method that selects key features uses the KS test to detect distribution changes between training and test sets.
Similarly,~\cite{ginart2022mldemon} applies KS test to drift detection and enhances it with periodic expert label queries after drift is detected, enabling more accurate performance monitoring.
\vspace{5pt}

\noindent \textbf{Chi-Square Test.}
This method is commonly used for categorical data, such as gender, hospital unit, or diagnosis code. The idea is to compare how often each category appears in two datasets. If the difference in frequencies is too large, the test signals a potential shift.
In \cite{merkow2023chexstray}, chi-square test is applied to DICOM metadata (\eg, gender) to detect categorical distribution shifts in imaging data monitoring.
Chi-square test is simple, fast, and interpretable, but limited to categorical features and cannot directly handle continuous variables.

\subsubsection{Sequential Statistical Methods}

These methods detect distributional shifts in streaming or temporally ordered data by continuously monitoring summary statistics or deviations from a baseline, making them well-suited for gradual drift detection in real-time settings.
\vspace{5pt}

\noindent \textbf{Cumulative Sum (CUSUM).}
CUSUM~\cite{romano2023fast} is a statistical monitoring method for detecting data drift by tracking small, incremental deviations from an expected value. When the cumulative sum exceeds a set threshold, it flags a potential shift. This approach is effective for catching subtle, gradual changes and is commonly used in quality control and performance monitoring.
The study~\cite{ganitidis2024sustaining} investigates AI performance degradation on two audio COVID-19 datasets using the CUSUM algorithm to monitor changes in the MMD statistic over time. By tracking the cumulative sum of MMD values across sliding data windows, it detects sustained deviations that signal shift.
\vspace{5pt}

\noindent \textbf{Statistical Process Control (SPC).}
SPC is a versatile and broadly applicable framework for monitoring healthcare data over time. Originally developed for manufacturing, SPC is now widely used in machine learning applications to detect shifts/anomalies in data distributions or model performance. It encompasses a range of statistical tools, including control charts, exponentially weighted moving averages (EWMA), and cumulative sum.
Zamzmi~\etal\cite{zamzmi2024out} applied SPC with CUSUM to monitor feature distribution shifts in radiology data, effectively detecting out-of-distribution cases across imaging modalities and demographic groups, demonstrating SPC's robustness in dynamic clinical settings.


\subsection{Machine Learning-based Methods}

\subsubsection{Supervised Methods}
These methods detect data shift by training a domain classifier (\eg, logistic regression) to distinguish between source and target samples~\cite{dreiseitl2022comparison}. The classifier's performance, measured by AUC, reflects distributional difference: AUC near 0.5 suggests no shift, while a high AUC indicates a clear separation. This approach, also known as \textbf{Classifier Two-Sample Test (C2ST)}~\cite{lopez-paz2017revisiting,kirchler2020two,pandeva2024evaluating}, is simple, flexible, and effective in high-dimensional data.

The study~\cite{Koch2024DistributionSD} applies this method to monitor an AI system developed for diabetic retinopathy. Their classifier detects shift related to image quality, co-morbidities, and demographics, outperforming traditional statistical tests but requiring larger training datasets. \cite{barrabes2024adversarial} uses an adversarial learning approach, training a discriminator to detect shifts between datasets.
Given that real-world data shift can emerge gradually, domain classifiers should ideally be updated in an online fashion to maintain reliable shift detection over time~\cite{jang2022sequential}.
\subsubsection{Unsupervised Methods}
These methods do not require labeled data, and often use \emph{autoencoders} trained to reconstruct inputs from a source distribution~\cite{berahmand2024autoencoders}. When applied to target data, a significant increase in reconstruction error (\eg, mean squared error) indicates data shift.
In addition to reconstruction error, these methods can monitor changes in the distribution of latent feature distributions to detect more subtle or structural shifts. For example, \cite{bobeda2023unsupervised} uses convolutional autoencoders to detect drift in breast cancer mammography data by analyzing both reconstruction error and latent space changes.

Unsupervised methods are well-suited to medical settings with labeled data are scarce.
However, their performance can be sensitive to model design, training quality, and data noise. Furthermore, they may fail to detect shifts that are not well captured by the reconstruction task.

\subsection{Feature-Based Methods}
These methods focus on detecting the changes in the relationship between input features and model predictions over time. Instead of monitoring raw data or outputs alone, they analyze shifts in feature contributions to the model's decision-making process, capturing changes in either the input distribution (\(p(x)\)) or the conditional distribution (\(p(y|x)\)). 
This is typically achieved using explainability tools like SHAP~\cite{SHAP,baptista2022relation} or by training auxiliary classifiers to identify which features are driving the observed shift~\cite{wang2024feature}.
The study~\cite{duckworth2021using} analyzes a model trained on pre-COVID emergency department data and observes performance degradation during COVID. By tracking SHAP values over time, they identify shifts in the importance of respiratory-related features, revealing clinically meaningful drift.
The study~\cite{barrabes2024adversarial} trains a binary classifier to distinguish source from target data and uses feature importance scores to identify variables driving the shift.

These methods are also very useful for root cause analysis, as they offer both detection and interpretability, which is critical in complex systems like healthcare~\cite{allgaier2023does}.
\subsection{Out-of-Distribution Detection}
While data shift detection tracks distribution changes, out-of-distribution (OOD) detection focuses on instance-level abnormalities, \ie, rare, unexpected, or extreme inputs that deviate from the training distribution~\cite{yang2024generalized}. 
OOD detection, also referred to as anomaly or outlier detection, plays a vital role in identifying high-risk cases that may trigger model failures before widespread drift occurs.

Traditional OOD detection methods use model confidence signals like Maximum Softmax Probability (MSP)~\cite{hendrycks2017a}, but often fail on high-dimensional, heterogeneous clinical data with poorly calibrated confidence scores.
Recent advances have introduced more robust approaches:
1) energy- and density-based methods~\cite{liu2020energy} that better capture data geometry and uncertainty;
2) contrastive and representation learning~\cite{sehwag2021ssd} that improve in–out separability in embedding space; and
3) diffusion-based methods~\cite{yang2024diffusion} that yield richer probabilistic estimates for complex or multimodal healthcare data.

Domain-specific methods have also emerged, such as NERO~\cite{chhetri2025nero}, which enables explainable OOD detection for gastrointestinal imaging, and MultiOOD~\cite{dong2024multiood}, which extends OOD evaluation across text, image, and tabular modalities. However, OOD benchmarking in healthcare remains limited, as general frameworks like OpenOOD~\cite{zhang2023openood} rarely capture real-world clinical variability.

OOD detection plays two key roles: 1) as a pre-filter to exclude extreme outliers before shift detection, and 2) as a standalone safeguard for identifying early or localized system failures. For instance, \cite{park2021reliable} shows that anomaly detection improves the reliability of healthcare AI systems by filtering out unreliable inputs and preventing erroneous predictions.

\section{Monitoring AI: Model Drift Detection}
\subsection{Performance-Based Methods}
These methods detect model drift by tracking changes in metrics such as accuracy, error rate, F1 score, or ROC-AUC using labeled data~\cite{kore2024empirical}. They compare performance across time windows or batches to identify significant degradation.

A classic example is \cite{gama2004learning}, which monitors online error rates and triggers drift alarms when errors exceed a threshold, prompting model retraining or recalibration.
This model-agnostic method was applied by Rotalinti~\cite{rotalinti2022detecting} to clinical prediction models using the Wilcoxon test on ROC-AUC over time, showing that even high-performing models can experience silent drift in dynamic healthcare settings.

While performance-based methods offer clear and direct insights into model health, their reliance on labeled data limits their applicability for real-time monitoring. In practice, they are more suitable for retrospective analysis or scheduled (\eg, periodic) performance audits.

\subsection{Model Output-Based Methods}
These methods detect drift by tracking shifts in predicted outputs (\eg, softmax probabilities or confidence scores) without relying on ground-truth labels~\cite{kore2024empirical,lipton2018detecting,rabanser2019failing}. Such outputs act as compact proxies for model behavior, enabling scalable, label-free, and model-agnostic monitoring, which is particularly valuable in healthcare, where labels are often delayed or unavailable. 

\subsubsection{Distance-Based Output Monitoring}
This approach uses the full \textbf{softmax output vector} as a proxy for model behavior, comparing its distribution between source and target data using statistical distance measures, without requiring labels.

A notable example is the Black Box Shift Detection (BBSD)~\cite{kore2024empirical}, which monitors predicted probabilities across 14 disease classes. By applying MMD to output distributions over time, BBSD detects shifts in unlabeled chest X-ray data, such as those introduced by COVID-19. Its sensitivity improves when combined with features from an autoencoder.
Similarly,~\cite{huggard2020detecting} uses predicted probabilities to detect model drift in medical triage systems.

These methods are especially effective in multi-class settings, capturing subtle class-specific shifts through the full output vector and offering deeper insight into model behavior.
\subsubsection{Score-Based Output Monitoring}
Score-based methods detect behavioral shifts by tracking scalar statistics derived from model outputs, such as maximum softmax probability, entropy, or energy scores from raw logits. These metrics aim to capture changes in model confidence or uncertainty that may indicate distributional drift, even without labels.

The study~\cite{hendrycks2017baseline} uses maximum softmax probability to flag prediction errors and distribution shifts. Evaluated on both vision and text datasets, the method shows that misclassified or OOD inputs tend to have lower softmax scores, making it a widely adopted baseline for OOD and drift detection.
Energy-based methods \cite{liu2020energy} extend this idea by computing a log-sum-exp energy score from logits, which captures overall activation levels and is sensitive to subtle uncertainties.

These methods are lightweight, interpretable, and suitable for real-time monitoring. However, their effectiveness may be limited in poorly calibrated models or when drift affects the distribution in ways not reflected by a single summary score.
\subsection{Auxiliary Model-Based Method}
These methods detect model drift by using a separate model to estimate prediction reliability, instead of relying solely on the primary model’s outputs (\eg, softmax scores). These auxiliary models learn external indicators of error or trustworthiness using either discriminative or generative approaches.

The study~\cite{amoukousequential} trains an auxiliary error estimator on labeled source data to predict instance-level errors. During deployment, the system monitors the rate of high-error predictions and raises an alert when the rate exceeds a predefined threshold.
In~\cite{che2021deep}, a conditional generative model (\eg, cVAE) is used to estimate $p(x | y)$, the likelihood of the input given its predicted label. A low likelihood implies potential unreliable and serves as a signal for detecting distribution shift.

These methods are modular and flexible, preserving the primary model's integrity while enabling auxiliary modules to be trained/adapted independently to dynamic environments. The drawback is that extra computation is typically needed.
\subsection{Calibration Drift Detection}
Calibration drift occurs when a model's predicted probabilities no longer align with actual outcome frequencies. For example, if a model predicts a class with 0.8 confidence, that prediction should be correct about 80\% of the time. In dynamic clinical settings, this alignment can degrade over time.

The study~\cite{davis2020detection} proposes a calibration drift detection system that uses dynamic calibration curves updated with true outcomes. An adaptive windowing algorithm monitors calibration error and flags drift when error exceeds a defined threshold, while identifying recent stable data for model recalibration.

\cite{shashikumar2023unsupervised} detects calibration drift without labels by summarizing patterns from new patient data (\eg, average risk scores) and using a regression model to estimate calibration quality. If adjustments are needed to meet a target metric, drift is inferred.

Because calibration drift can silently compromise clinical decision-making, ongoing and monitoring and timely recalibration are essential for safe AI deployment in healthcare.

\section{Root Cause Analysis}

In high-stakes fields like healthcare, merely detecting performance degradation is insufficient, understanding its underlying cause is essential for safe and effective model adaptation. Root Cause Analysis (RCA) offers a structured approach to identifying the reasons behind model failure, enabling targeted, diagnosis-driven interventions.

Traditional correction methods, such as domain adaptation, are often reason-agnostic and apply the same fix to all issues. This approach is analogous to administering treatment without a proper diagnosis, risking ineffective or harmful outcomes. RCA bridges this gap between drift detection and remediation, guiding more precise and informed responses.
As such, RCA is increasingly recognized as a critical component in building trustworthy and robust AI monitoring systems.

\subsection{LLM-Based Diagnostic Method}
Rauba~\etal\cite{rauba2025self} introduce H-LLM, the first self-healing AI system that uses large language models (LLMs) to autonomously diagnose root causes of model degradation and recommend targeted adaptation strategies. In evaluations across healthcare and finance, H-LLM outperforms traditional methods by reducing false-positive drift detections and improving adaptation outcomes.

The study highlights two key points: (1) root cause analysis is essential for effective, diagnosis-driven adaptation, and (2) LLMs show strong potential in enabling intelligent, self-monitoring AI systems. By integrating reason-aware diagnostics, AI models can transition from reactive correction to proactive, self-healing behavior, enhancing reliability and resilience in deployment.

\subsection{Shift Type Disentanglement Method}
Accurate identification of the specific type of distribution shift is essential for effective RCA. The study~\cite{roschewitz2024automatic} proposes a framework that differentiates between label shift, covariate shift, and hybrid shift. The approach uses model output distributions to detect label shifts and leverages feature-based analysis via self-supervised encoders to detect covariate shifts.
The combined ``Duo detector" which integrates both techniques, achieves superior performance across various shift types. Evaluated on large-scale datasets from chest X-rays, mammography, and retinal fundus images, the method demonstrates strong generalizability.

The key takeaway is that each shift type necessitates tailored detection and correction strategies. For example, label shift often requires output reweighting, whereas covariate shift is better addressed through feature-level adjustments.

\subsection{Causal Inference-Based Method}
Causal inference~\cite{yao2021survey,prosperi2020causal} offers an interpretable framework for tracing model degradation to upstream data or policy shifts. Using causal graphs, it reveals how variables influence each other, enabling AI systems to diagnose why a shift occurred and identify corrective actions, rather than passively adapting.
\cite{subbaswamy2020development} applies this approach to a pneumonia risk model that fails when transferred to a new hospital. The causal graph revealed that the model's performance drop is linked to differences in ICU admission policies.

Causal root cause analysis moves beyond correlation, enabling targeted interventions (whether adjusting the model, data pipeline, or clinical workflow) to restore model stability.

\subsection{Model-Centric Analysis of Internal Dynamics}
Another perspective on RCA involves examining the internal parameters of neural networks. For example, WeightWatcher~\cite{WeightWatcher} evaluates the spectral properties of weight matrices across layers in pretrained deep networks. Using norm-based and power-law-based metrics grounded in heavy-tailed self-regularization theory, WeightWatcher can detect when a model exhibits capacity loss, over-parameterization, or under-training. Such approaches highlight the value of incorporating layer-wise diagnostics into RCA, offering a model-centric complement to data-centric or causal inference methods.

\subsection{Illustrative Example}
Consider a clinical decision support model for pneumonia detection from chest X-rays. Over time, its accuracy drops when deployed in a new hospital. An RCA pipeline would proceed as follows:
\begin{enumerate}
    \item Detection: Model monitoring identifies a significant performance drop for a specific patient subgroup (\eg, elderly patients).
    \item Data-level analysis: Data monitoring reveals a shift in X-ray device type and contrast distribution compared to the training site.
    \item Model-level analysis: Inspection of internal activations shows reduced feature sensitivity, \ie, a weakened correlation between intermediate layer features and the classification output, in mid-level convolutional layers.
    \item Correction: Domain adaptation or calibration adjustment restores performance.
\end{enumerate}

This example illustrates how RCA integrates data- and model-level analyses to identify the underlying cause of degradation and guide corrective action, which is essential for maintaining safe and reliable medical AI systems.

\begin{figure}[t]
\setlength{\belowcaptionskip}{-2pt}
\setlength{\abovecaptionskip}{0pt}
\setlength{\abovedisplayskip}{0pt}
\setlength{\belowdisplayskip}{0pt}
\center
 \includegraphics[width= 1.0\linewidth]{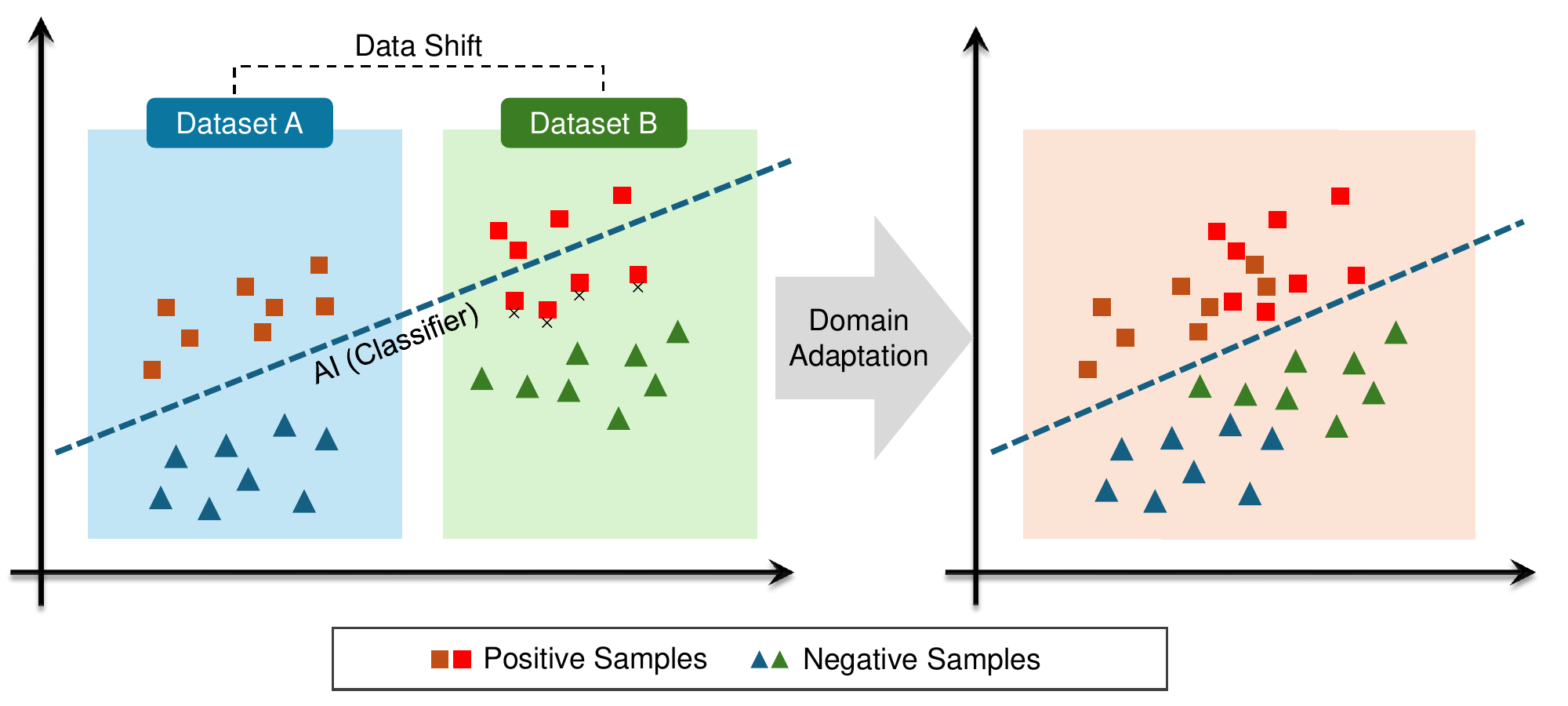}
 \caption{Illustration of domain adaptation which can correct AI performance degradation caused by data shift.} 
 \label{Adaptation}
\end{figure}


\begin{table*}[t]

\centering
\caption{Quantitative comparison of representative methods across the Detection-Diagnosis-Correction framework.}
\label{tab:ddc_summary}
\small
\begin{tabular}{l l l c}
\toprule
\textbf{Stage} & \textbf{Method Family (Examples)} & \textbf{Prerequisites} & \textbf{Cost} \\
\midrule
\textbf{Detection} 
& Statistical distance measures (MMD, JS) 
& Reference data, feature representations
& Medium-High \\
& Statistical test (KL test, Chi-Square Test) 
& Reference data, feature/output distributions 
& Low \\
& Supervised methods (Domain classifier)
& Access to cross-site data, model training
& Medium-High \\
& Unsupervised methods (Autoencoder)
& Access to source data, model training
& Medium \\
& Feature-based methods (SHAP)
& Access to model internals/predictions, labeled samples
& High \\
& Out-of-Distribution Detection (MSP)
& Reference data, model outputs or feature embeddings
& Medium \\
& Performance analysis (Accuracy)
& Labeled target data
& Low \\

\midrule
\textbf{Diagnosis} 
& LLM-based diagnosis
& Access to LLM resources, reference data
& High \\
& Shift type disentanglement
& Reference data, labeled samples, model outputs 
& Medium-High \\
& Causal inference
& Labeled data with causal variables, model access 
& High \\

\midrule
\textbf{Correction} 
& Unsupervised domain adaptation 
& Source and target data, model access 
& High \\
& Test‐time adaptation
& Model access, unlabeled stream 
& Medium-High \\
& Re-training and Fine-Tuning 
& Labeled data, model training
& High \\
& Continual learning 
& Sequential data, model update access
& Medium-High \\
\bottomrule
\end{tabular}
\vspace{3pt}

\end{table*}


\section{Correcting AI Performance Degradation}

\subsection{Domain Adaptation}

Domain adaptation~\cite{guan2021domain}, as shown in Fig~\ref{Adaptation}, is a technique designed to reduce the performance gap of machine learning models when deployed in a target domain (\eg, a new hospital, population, or device) whose data distribution differs from that of the source domain used during training.

\subsubsection{Unsupervised Domain Adaptation (UDA)}
It assumes access to labeled data in the source domain and unlabeled data in the target domain~\cite{wilson2020survey}. The goal is to align the latent representations across domains without requiring labels from the target side.

In the study~\cite{ganitidis2024sustaining}, UDA is implemented to correct for data drift in COVID-19 detection models using audio data. After drift detection, UDA minimizes the MMD between the source (development) and target (post-development) data distributions.
This process aligns the feature representations of the two domains.
Experimental result validates UDA's potential to maintain robust performance in evolving healthcare scenarios. 
\subsubsection{Test-Time Domain Adaptation (TTDA)}
TTDA~\cite{liang2025comprehensive} refers to the methods that adapt models on-the-fly during inference using only unlabeled target data observed at test time. Unlike UDA, TTDA does not require access to target data during training and operates under more constrained but realistic conditions.

TTDA has been applied in cross-site medical image classification and segmentation tasks, demonstrating adaptability to distribution shifts, particularly covariate shifts~\cite{valanarasu2024fly,yang2022dltta,wen2024denoising,aleem2024test}.
Recently, TTDA has also been extended to LLMs in biomedical NLP. \cite{shi2024medadapter} introduces TTDA to improve LLM performance on tasks such as biomedical question answering, showing that even large pretrained models can benefit from dynamic adjustment to target-domain inputs at inference time.


\subsection{Re-training and Fine-Tuning }
\subsubsection{Re-Training}
It involves updating a model with new data to counteract performance degradation and maintain accuracy and reliability as input distributions or outcome patterns shift.

The study~\cite{rahmani2023assessing} examines the effects of event-driven shifts (\eg, during COVID-19) on sepsis prediction models. Simulation results show that retrained models, especially XGBoost, outperformed static models, highlighting the importance of regular retraining to maintain performance.
\cite{adam2020hidden} explores how feedback loops in healthcare can degrade models over time. Using MIMIC-IV data, they show that repeated updates may increase false positive rates, even when AUC appears stable. Their findings indicate that retraining on the full historical data is more robust than incremental updates using only recent data.

Despite its effectiveness, retraining is resource-intensive, requiring quality labels and full validation cycles, which may limit its feasibility in fast-paced clinical settings.
\subsubsection{Fine-Tuning (Transfer Learning)}
It adapts a pretrained AI model to a new domain by updating all or part of its parameters using some target data. It is particularly effective when source and target domains share structural similarities but differ in characteristics (\eg, patient demographics or imaging protocols).

The study~\cite{subasri2025detecting} fine-tunes an in-hospital mortality prediction model originally trained at one hospital using data from different target hospitals. The results show that transfer learning can mitigate performance degradation and help maintain model performance across diverse healthcare settings.

Fine-tuning offers a cost-effective and flexible correction method, especially when labeled data are scarce but pretrained models offer transferable features. 

\begin{table*}[t]

\caption{Datasets for Medical AI Performance Monitoring Research.}

\centering
\renewcommand{\arraystretch}{1.5} 
\setlength{\tabcolsep}{22.5pt} 
\renewcommand{\arraystretch}{1.5}

\begin{tabular}{l l l l l l } 
\toprule

\textbf{Data Type}  & \textbf{Dataset}     &\textbf{Country}   & \textbf{Time Span}   
                    & \textbf{Samples}   & \textbf{Studies} \\ 
\hline

\multirow{2}{*}{EHR} 
&MIMIC-III   &USA   &2001–2012    &$\sim$50,000     &\cite{MIMIC3,schrouff2022diagnosing}  \\

\cline{2-6}

&MIMIC-IV    &USA   &1996–2019    &$\sim$380,000     &\cite{MIMIC4,adam2020hidden} \\ 

\hline

\multirow{3}{*}{Imaging}      
&CheXpert    &USA   &2002--2017   &$\sim$220,000     &\cite{irvin2019chexpert,merkow2023chexstray,guo2023medshift}  \\ 

\cline{2-6}

&MIMIC-CXR    &USA   &2011--2016   &$\sim$380,000     &\cite{johnson2019mimic,guo2023medshift}  \\   

\cline{2-6}

 &PadChest  &Spain   &2009--2017   &$\sim$160,000   &\cite{bustos2020padchest,merkow2023chexstray} \\                      

\cline{2-6}

 &CAMERON  &Netherlands   &2006--2016   &$\sim$450,000   &\cite{roschewitz2023automatic,litjens20181399} \\  

\hline
\multirow{2}{*}{Audio}  &Coswara dataset  &World  &2020--2022   &$\sim$23,000   
                        &\cite{bhattacharya2023coswara,ganitidis2024sustaining}    \\ 
                        \cline{2-6}
                        &COVID-19 Sounds  &World  &2020--2021  &$\sim$50,000     &\cite{xia2021covid,ganitidis2024sustaining}    \\

\hline
EEG    &TUH EEG Corpus    &USA   &2002–    &$\sim$60,000     &\cite{obeid2016temple,bhaskhar2023explainable} \\

\hline
\end{tabular}

\label{tab:datasets}
\end{table*}

\subsection{Continual Learning (CL)}
CL~\cite{bruno2025continual,liu2017lifelong} enables models to adapt to new data while retaining prior knowledge, offering dynamic updates suited to evolving healthcare environments (\eg, disease patterns, practices, and demographics).

\cite{lee2020clinical} highlights the potential of CL in diagnostic imaging, prediction, and treatment planning, while also noting challenges such as catastrophic forgetting, computational overhead, and regulatory concerns. 
\cite{subasri2025detecting} applies drift-triggered CL in a hospital setting, detecting temporal shifts and retraining on updated data. This approach successfully maintains model performance under changing clinical conditions.
\cite{chi2022novel} proposes a lifelong learning framework using knowledge distillation to mitigate calibration drift, outperforming standard retraining on four cancer datasets by maintaining both calibration and discrimination over time.

These approaches are essential for real-world deployment but require careful handling of memory, stability, and update frequency to prevent overfitting or degradation.

\subsection{Calibration Correction}
Calibration correction~\cite{guo2017calibration} adjusts a model’s predicted probabilities to better reflect true outcome rates. Unlike retraining methods, it modifies outputs directly, maintaining accurate risk estimates. This is important in healthcare, where decisions often depend on calibrated thresholds.
Su~\etal\cite{su2018review} categorize calibration updates into 1) coefficient adjustment, 2) meta-model combination, and 3) dynamic updating with new data. Using cardiac surgery data, they show that adaptive updates improve reliability in clinical settings.

Recent studies explore label-free approaches. \cite{shashikumar2023unsupervised} uses cohort-level features to detect and correct calibration shifts without labels. Similarly, \cite{roschewitz2023automatic} aligns prediction distributions across imaging datasets to mitigate acquisition shifts, preserving performance in tasks like breast cancer detection.

Calibration methods are lightweight and effective when model ranking remains reliable but confidence levels drift. However, they cannot address changes in decision boundaries or underlying concepts.

%
\subsection{Adaptive Update Strategies}
Update strategies define when and how to apply corrections for performance degradation, offering a broader framework for managing shifts of varying scale and frequency. Effective strategies balance timely intervention with model stability.
Key questions include: 1) \textbf{When to update?} Only if the shift significantly impacts performance; 2) \textbf{How to update?} Options include recalibration, partial fine-tuning, or full retraining. 3) \textbf{How often to update?} Avoid excessive or delayed updates to maintain reliability.

In~\cite{janssen2008updating}, five strategies are evaluated for a clinical prediction model. Simple recalibration often performs as well as more complex methods. Similarly,~\cite{davis2019nonparametric} proposes a bootstrap-based adaptive strategy that selects update types based on degradation severity. Tested on several clinical models, it finds recalibration sufficient in most cases, with full retraining needed only for major shifts.
%
\vspace{8pt}

To summarize the above methods with the qualitative discussion, Table \ref{tab:ddc_summary} presents representative approaches within the Detection-Diagnosis-Correction framework, highlighting their main prerequisites and relative computational cost.

\section{Datasets, Tools and Benchmarks}
\subsection{Datasets}

Medical AI monitoring remains challenging due to the scarcity of publicly available datasets and standardized benchmarks, as most studies rely on proprietary hospital data. Advancing the field requires open datasets that capture both ``natural shifts" (\eg, those arising from real-world events like the COVID-19 pandemic) and ``synthetic shifts" (\eg, engineered changes in demographics or image quality) to evaluate model robustness.
The COVID-19 pandemic highlights the need to monitor medical AI. For instance, the University of Michigan Hospital deactivated a sepsis-alerting model in April 2020 after demographic shifts triggered spurious alerts~\cite{finlayson2021clinician}. Such shifts can disrupt clinical variable relationships, degrade model performance, and increase patient risks. 

Table~\ref{tab:datasets} summarizes several representative datasets across different modalities that have been used to study medical AI performance degradation detection.
\subsection{Performance Monitoring and Correction Tools}
With growing attention to AI performance monitoring post-deployment, several commercial and open-source tools have been developed to support this task.

Prominent commercial solutions include Amazon SageMaker Model Monitor~\cite{nigenda2022amazon}, Google Vertex AI Model Monitoring~\cite{taly2021monitoring}, and Microsoft Azure MLOps~\cite{edwards2022mlops}, all offering integrated tools for tracking model behavior in production.

Notable open-source alternatives include Evidently AI~\cite{evidentlyai} and Deepchecks~\cite{Deepchecks}, which provide modular components for detecting data drift, monitoring performance, and generating reports. Other useful tools include TorchDrift~\cite{torchdrift} and Frouros~\cite{frouros}, both of which offer libraries for statistical drift detection and monitoring.

DomainATM~\cite{guan2023domainatm} is a toolbox for domain adaptation in medical data analysis. Developed in MATLAB with a user-friendly graphical interface, it includes a suite of widely used adaptation algorithms tailored for medical data analysis tasks. 
\subsection{Benchmarks}

\subsubsection{Stanford WILDS/Wild-Time Benchmarks}
The Stanford WILDS benchmark~\cite{koh2021wilds}\footnote{\url{https://wilds.stanford.edu/}} evaluates model robustness under real-world distribution shifts across domains like healthcare and ecology. 
Results show significant performance drops from in- to out-of-distribution settings, underscoring the challenge of robust deployment.
Extending this, Wild-Time~\cite{yao2022wild}\footnote{\url{https://wild-time.github.io}} focuses on temporal drift using five real-world datasets, including MIMIC-IV for clinical prediction tasks.
Wild-Time reports around 20\% performance drops of AI models over time, highlighting the difficulty of achieving temporal robustness.

\subsubsection{TableShift Benchmark}
TableShift~\cite{gardner2024benchmarking}\footnote{\url{https://tableshift.org}} benchmarks distribution-shift robustness in tabular data, an area long lacking standardized evaluation compared to text and vision. This is critical in healthcare, where structured EHR data are common. It includes tasks such as ICU mortality, length of stay, and sepsis prediction, supports models like FT-Transformer~\cite{gorishniy2021revisiting} and Tabular ResNet~\cite{borisov2022deep}.
\subsubsection{Concept Drift Detection Benchmark}
In the Large-Scale Comparison of Concept Drift Detectors~\cite{barros2018large}, 14 drift detection methods are evaluated on synthetic datasets with abrupt and gradual shifts, using Naïve Bayes and Hoeffding Tree as the two basic classifiers.
It evaluates both classification performance and drift detection accuracy, and analyze their correlation. One interesting finding is that precise detection of all drifts may not always lead to better outcomes. 
%
\subsubsection{Discussion: Benchmarking and the Evaluation Gap}
While several public datasets support data shift and OOD research (\eg, WILDS, MedShift, OpenOOD), most benchmarks remain general-purpose and do not fully capture the clinical complexity of healthcare settings. As recently emphasized by~\cite{mahmood2025benchmarking}, biomedical machine learning faces a ``benchmarking crisis", where performance is often reported without rigorous or standardized evaluation across institutions, populations, and acquisition settings. This limitation extends to AI monitoring, where few benchmark suites simulate longitudinal or multi-site deployment. Establishing standardized, clinically grounded benchmarks for AI monitoring is therefore an urgent priority for enabling fair, reproducible, and thorough evaluation.
\section{Challenges and Future Directions}\label{Discussion}
\subsection{Major Challenges}
\subsubsection{Missing or Delayed Ground-Truth Labels for AI Monitoring}

Real-time monitoring enables early detection of AI performance degradation, allowing timely clinical response. However, ground-truth labels, the gold standard for evaluation, are often delayed or infeasible in clinical settings. Label generation typically requires expert annotation (\eg, chest X-ray annotations for pulmonary diagnoses) or long-term outcomes (\eg, mortality, disease progression). Such delays hinder timely assessment and correction, especially for predictive and diagnostic models.
\subsubsection{Human-Centric Requirement Monitoring}

Medical AI monitoring has traditionally emphasized performance metrics such as diagnostic accuracy. However, recent research underscores the need to address human-centric concerns such as privacy, fairness, and alignment with human values~\cite{naveed2024towards}. Performance metrics alone are insufficient. For example, a system may show stable accuracy overall while underperforming for certain ethnic groups. To ensure equitable and trustworthy AI, monitoring must also detect and address such disparities. Future efforts should incorporate tools like bias and privacy detectors to complement performance monitors and support both technical reliability and societal accountability.
\subsubsection{Stability-Plasticity Dilemma}

A key challenge in updating AI models after degradation is the \textbf{stability-plasticity dilemma}~\cite{mermillod2013stability}, the trade-off between retaining prior knowledge (stability) and adapting to new information (plasticity). Stability ensures consistent clinical support but insufficient updates cause obsolescence.
Plasticity enables adaptation to evolving medical knowledge, disease patterns, and clinical guidelines, but frequent updates risk catastrophic forgetting~\cite{kirkpatrick2017overcoming,kemker2018measuring}. Balancing these forces is difficult: excessive update destabilizes models, while infrequent updates cause performance decay. Effective adaptive strategies are crucial to keep AI systems both reliable and relevant.

\subsubsection{System Complexity and Maintenance Challenges}

Real-world AI systems are increasingly complex, featuring feedback loops, dynamic data pipelines, and interdependent modules. Unlike rule-based software, their data-driven nature makes them sensitive to input shifts and operational changes. This complexity heightens the risk of silent degradation: feedback loops can amplify bias, and changes in one component may destabilize others, accumulating technical debt~\cite{2015Hidden}. These challenges are further magnified in large models such as LLMs and VLMs, whose black-box architectures and emergent behaviors make interpretation and monitoring difficult~\cite{yampolskiy2024monitorability}.

\subsubsection{Prompt-Injection and Retrieval Drift in LLM-Based Systems}
For LLM-based medical systems, new forms of drift emerge beyond traditional data or model shifts and require specialized monitoring strategies. Prompt injection can be viewed as an input-level anomaly in which malicious or unintended instructions alter model behavior. Such anomalies can be detected by monitoring deviations in prompt structure, intent, or embedding similarity against validated templates. Retrieval drift refers to the gradual evolution of external knowledge sources in retrieval-augmented generation (RAG) pipelines, leading to inconsistent or outdated information. It can be monitored through periodic validation of retrieved content. Both forms of drift introduce new challenges that call for further research.

\subsubsection{Multimodal AI Monitoring and Missing Modality Challenges}
As medical AI systems increasingly integrate diverse data types (\eg, imaging, EHRs, genomics, and text), monitoring multimodal models introduces additional complexity. Unlike unimodal systems, multimodal AI must maintain consistency across heterogeneous input spaces, making it sensitive to cross-modal drift (\eg, distribution shifts in one modality that alter joint representations) and missing modality scenarios that frequently occur in real-world practice. Recent work, such as MultiOOD~\cite{dong2024multiood}, highlights scalable OOD detection across multiple modalities. Addressing such challenges requires monitoring frameworks that jointly assess modality-specific and cross-modal reliability, with correction strategies that adapt to missing or degraded modalities at inference.

\subsection{Future Directions}
\subsubsection{Label-Free Estimation}

In medical AI, ground-truth labels are often delayed or costly, limiting real-time monitoring. Label-free performance estimation offers an alternative, enabling models to self-assess without labeled target data.

One approach is using surrogate or pseudo labels from a reference model or the model itself. While useful, this risks propagating bias if the labels are inaccurate or unstable.
Another technique relies on model confidence scores, such as softmax outputs. For example, Average Thresholded Confidence (ATC)~\cite{gargleveraging} learns a confidence threshold from labeled source data to estimate accuracy on unlabeled target data.
A more principled method uses Optimal Transport (OT) to detect shifts between training and current data. In~\cite{koebler2024incremental}, OT quantifies feature distribution divergence to estimate performance degradation, optionally triggering active labeling when uncertainty is high. Similarly,~\cite{koebler2023towards} applies OT to input features for label-free performance estimation.

\subsubsection{Benign vs. Harmful Shift}
Not all distribution shifts cause significant performance degradation, so distinguishing between benign and harmful shifts is essential. Benign shifts have minimal impact and usually require no action, while harmful shifts degrade performance and demand intervention. Misidentifying benign shifts can lead to false alarms, inefficiencies, higher costs, and even system deactivation~\cite{finlayson2021clinician}.

Differentiating shift types improves AI system robustness by reducing false positives and focusing resources on meaningful interventions. However, most research emphasizes shift detection without assessing performance impact. Recent studies~\cite{podkopaev2022tracking} have begun addressing this by categorizing shift severity, offering a more practical approach for managing AI in dynamic environments.
\subsubsection{Root Cause Analysis}

Most correction methods are “reason-agnostic”, applying predefined corrections upon detecting performance degradation without examining its cause. This blind approach can lead to ineffective or even harmful updates, especially in healthcare, where it risks incorrect diagnoses or system failures.

To enhance reliability, medical AI should prioritize root cause analysis. Understanding the source of degradation (whether from data shifts, label noise, sensor errors, or training bias) enables targeted, evidence-based corrections instead of generic responses. This improves model robustness, transparency, and clinician trust. Recent work~\cite{rauba2025self} has begun leveraging LLMs for root cause analysis, harnessing their reasoning capabilities for more intelligent system updates.
\subsubsection{Leveraging Synthetic Data for AI Monitoring}

AI performance degradation often results from training data limitations, especially selection bias. For example, datasets from a single hospital may overrepresent specific groups, reducing model generalizability. Scarce and costly medical annotations also limit evaluation across diverse shift scenarios.

Synthetic data offer a promising solution~\cite{chen2021synthetic}. They can increase diversity, represent rare cases, and simulate different data shifts~\cite{piano2022detecting}, helping assess model behavior under varied conditions.
Simulated shifts also allow stress-testing of shift detection systems before deployment, aiding in identifying weaknesses and improving monitoring tools.
However, ensuring clinical realism remains a challenge. Future work should develop metrics that evaluate not just accuracy but also generalizability and safety when using synthetic data.

\subsubsection{Monitoring Large Language and Vision-Language Models}

LLMs and VLMs are increasingly used in healthcare, but their massive scale and black-box nature raise concerns about safety and reliability in clinical settings.
Recent studies reveal risks of behavioral drift. For example,~\cite{chen2024chatgpt} finds that GPT-4's instruction-following ability declined over time, highlighting the need for ongoing monitoring.
Performance degradation in LLMs/VLMs may result from knowledge staleness, hallucination, catastrophic forgetting, or few-shot performance decay. Empirical evidence~\cite{payne2024performance,bhayana2023performance,jeong2024limited} shows that GPT-4's diagnostic accuracy dropped within months across both text and image tasks.
As these models play a growing role in clinical decision-making, monitoring their performance is a critical and promising research direction.
\section{Conclusion}\label{sec13}

This survey presents a comprehensive framework for understanding and addressing AI performance degradation in healthcare, covering data and model drift detection, root cause analysis, and correction strategies. We also review representative open-source datasets, tools, and benchmarks that support medical AI monitoring. We highlight key challenges, such as delayed labels, fairness, the stability-plasticity trade-off, system complexity and multimodality, and emphasize future directions including label-free performance estimation, root cause analysis, synthetic data use, and continuous monitoring of large-scale models to ensure safety and reliability. 


\section*{Acknowledgements}
This work was supported by the National Library of Medicine under Grant No. 1R01LM014239.


\section*{References}
\vspace{-15pt}
\bibliographystyle{IEEEtran}
\bibliography{refs}

\end{document}